\ifdefined\pdfobjcompresslevel \pdfobjcompresslevel=0 \fi

\PassOptionsToPackage{table}{xcolor}

\documentclass[]{alaya}
\usepackage{makecell}
\usepackage{wrapfig}
\usepackage{tabularx}
\usepackage{textcomp}
\usepackage{stfloats}
\usepackage{url}
\usepackage{verbatim}
\usepackage{titlesec}
\usepackage{adjustbox}
\usepackage{multirow}
\usepackage{pifont}
\usepackage[sc]{mathpazo}
\usepackage{tikz}
\usepackage{comment}
\usepackage{amsmath,amssymb}
\usepackage{colortbl}
\usepackage{natbib}
\usepackage{color}
\usepackage{booktabs}
\usepackage{hyperref}
\usepackage{graphicx}
\usepackage{multirow}
\RequirePackage{xspace}
\makeatletter
\DeclareRobustCommand\onedot{\futurelet\@let@token\@onedot}
\def\@onedot{\ifx\@let@token.\else.\null\fi\xspace}
\usepackage[most]{tcolorbox}
\usepackage{xcolor}
\usepackage{array}
\usepackage{tabularx}
\usepackage{siunitx}
\usepackage{makecell}
\usepackage[table]{xcolor}
\definecolor{headerpurple}{HTML}{d8d2fc}
\definecolor{rowgray}{gray}{0.95}

\makeatother

\definecolor{adptorange}{RGB}{248, 205, 172}
\definecolor{cmpblue}{RGB}{189, 215, 238}

\definecolor{our_red}{RGB}{232,157,160}
\definecolor{our_blue}{RGB}{136,206,230}
\definecolor{our_orange}{RGB}{246,200,168}
\definecolor{our_green}{RGB}{178,211,164}

\definecolor{attn_code0}{RGB}{247,215,200}
\definecolor{attn_code1}{RGB}{238,169,139}
\definecolor{mlp_code0}{RGB}{204,201,221}
\definecolor{mlp_code1}{RGB}{102,95,153}
\definecolor{mygray}{HTML}{f0f0f0}

\definecolor{token_blue}{RGB}{84, 120, 140}

\usepackage{pifont}
\usepackage{bbding}
\usepackage{fontawesome}
\usepackage{xspace}

\usepackage{float}

\newlength\savewidth

\newcolumntype{x}[1]{>{\centering\arraybackslash}p{#1pt}}
\newcolumntype{y}[1]{>{\raggedright\arraybackslash}p{#1pt}}
\newcolumntype{z}[1]{>{\raggedleft\arraybackslash}p{#1pt}}

\renewcommand{\paragraph}[1]{\vspace{1mm}\noindent\textbf{#1}}
\usepackage{colortbl}
\usepackage{xcolor}
\usepackage{wrapfig}

\renewcommand{\paragraph}[1]{\vspace{1.25mm}\noindent\textbf{#1}}

\usepackage{algorithm}
\usepackage{listings}

\definecolor{codeblue}{rgb}{0.25, 0.5, 0.5}
\definecolor{codekw}{rgb}{0.35, 0.35, 0.75}
\lstdefinestyle{Pytorch}{
    language = Python,
    backgroundcolor = \color{white},
    basicstyle = \fontsize{9pt}{8pt}\selectfont\ttfamily\bfseries,
    columns = fullflexible,
    aboveskip=1pt,
    belowskip=1pt,
    breaklines = true,
    captionpos = b,
    commentstyle = \color{codeblue},
    keywordstyle = \color{codekw},
}

\definecolor{green}{HTML}{009000}
\definecolor{red}{HTML}{ea4335}

\title{\raisebox{-0.36\height}{\includegraphics[height=3.0em]{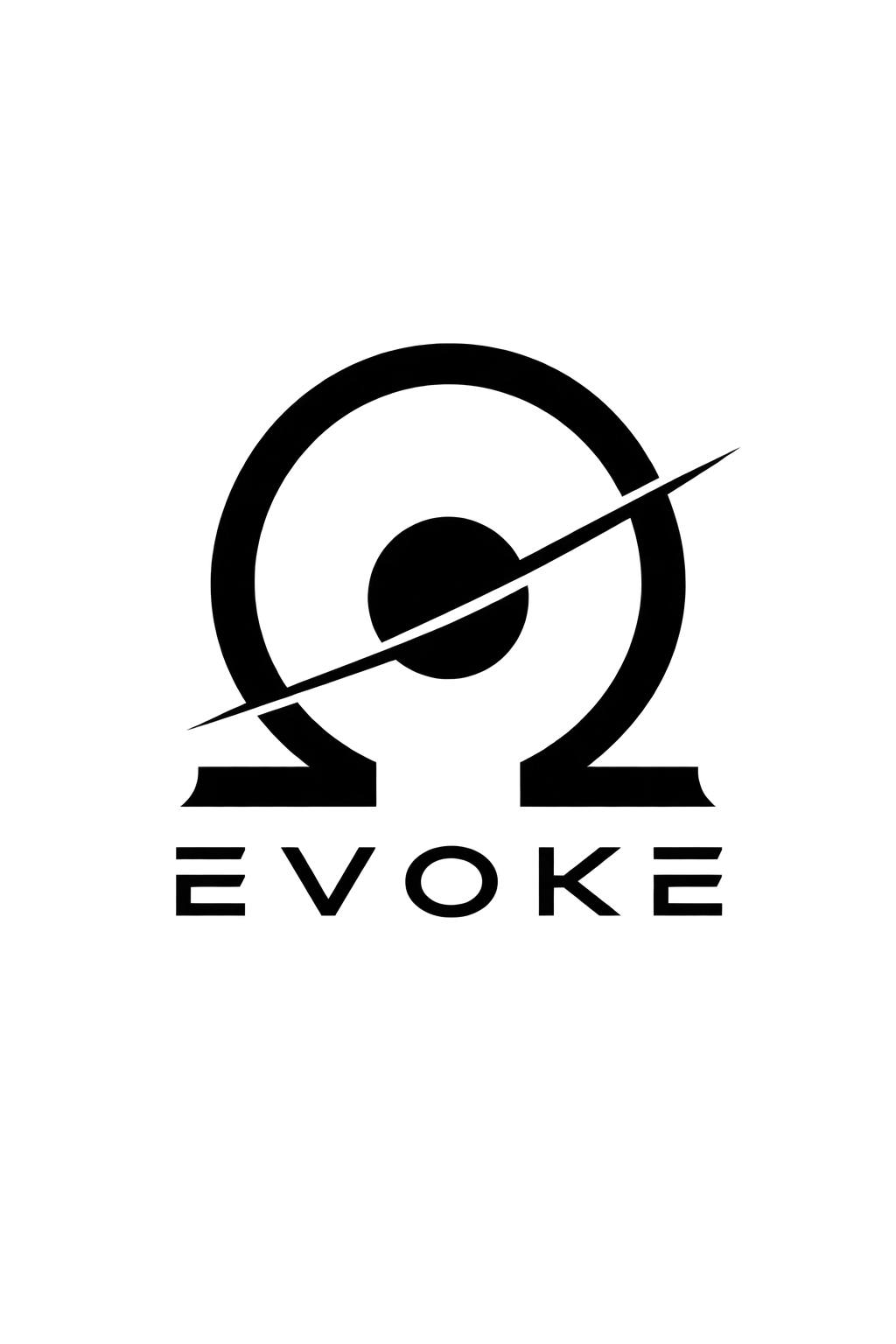}}\hspace{0.32em}Alaya-EVOKE: From Linear-Scaling Supervision to Endless World}

\author[1,2,3,*]{Yuanyang Yin}
\author[2,*]{Gongxuan Wang}
\author[3]{Yifan Zhan}
\author[3,\dagger]{Chuanhao Li}
\author[3,2, \dagger \ddagger]{Kaipeng Zhang}
\author[1,\ddagger]{Feng Zhao}

\affiliation[1]{MoE Key Lab of BIPC, USTC}
\affiliation[2]{Shanghai Innovation Institute}
\affiliation[3]{Alaya Lab}

\contribution[*]{Equal contribution, work done during internship at Alaya Lab}
\contribution[\dagger]{Project lead}
\contribution[\ddagger]{Corresponding author}

\correspondence{\email{yyyin@mail.ustc.edu.cn}}
\project{\url{https://evoke-world.github.io/Evoke/}}
\code{\url{https://github.com/SII-YuanyangYin/Evoke}}

\date{\today}

\abstract{
Interactive world models must simultaneously support persistent memory, responsive user interaction, and long-horizon generation, yet these requirements place conflicting demands on the underlying model. Maintaining history in the denoiser context or key-value cache incurs growing cost over time, forcing a trade-off between session length and retained memory, while low-latency interaction typically relies on few-step generation whose capabilities are ultimately bounded by its teacher. Alaya-EVOKE (Evoke) addresses both limitations by externalizing persistent world state and redesigning the teacher for long-horizon interactive generation. Scene geometry is maintained in an external, camera-indexed world state bank, from which only information relevant to the current view is retrieved, keeping the denoiser context bounded as the session grows. Rather than treating the teacher as a fixed high-quality generator, the teacher is explicitly designed for long-horizon supervision. Its sparse attention scheme combines chunk-wise grouping, retrieval of selected distant frames, and a linear-attention global state, yielding linear growth in activation memory and computation while enabling supervision over long temporal horizons. Such supervision exposes content drift that remains locally plausible within short windows, while per-chunk conditioning enables prompt changes and event control throughout the sequence. A 30-second long-horizon distribution-matching objective, applied under self-forced rollouts, transfers both capabilities to a three-step student that uses no classifier-free guidance (CFG), improving resistance to long-term content drift while preserving responsive conditioning. With bounded context and recurrent external memory, Evoke supports open-ended, continuously evolving generation; on a single H200 at $384\times640$, each $1.5$\,s chunk is generated in $2.11$\,s. As a three-step world model, Evoke achieves state-of-the-art performance on WBench while remaining competitive in visual quality on VBench-Long and VBench-2.0.
}

\begin{document}
\maketitle

\section{Introduction}

\begin{figure}[tb]
  \centering
  \includegraphics[width=\linewidth]{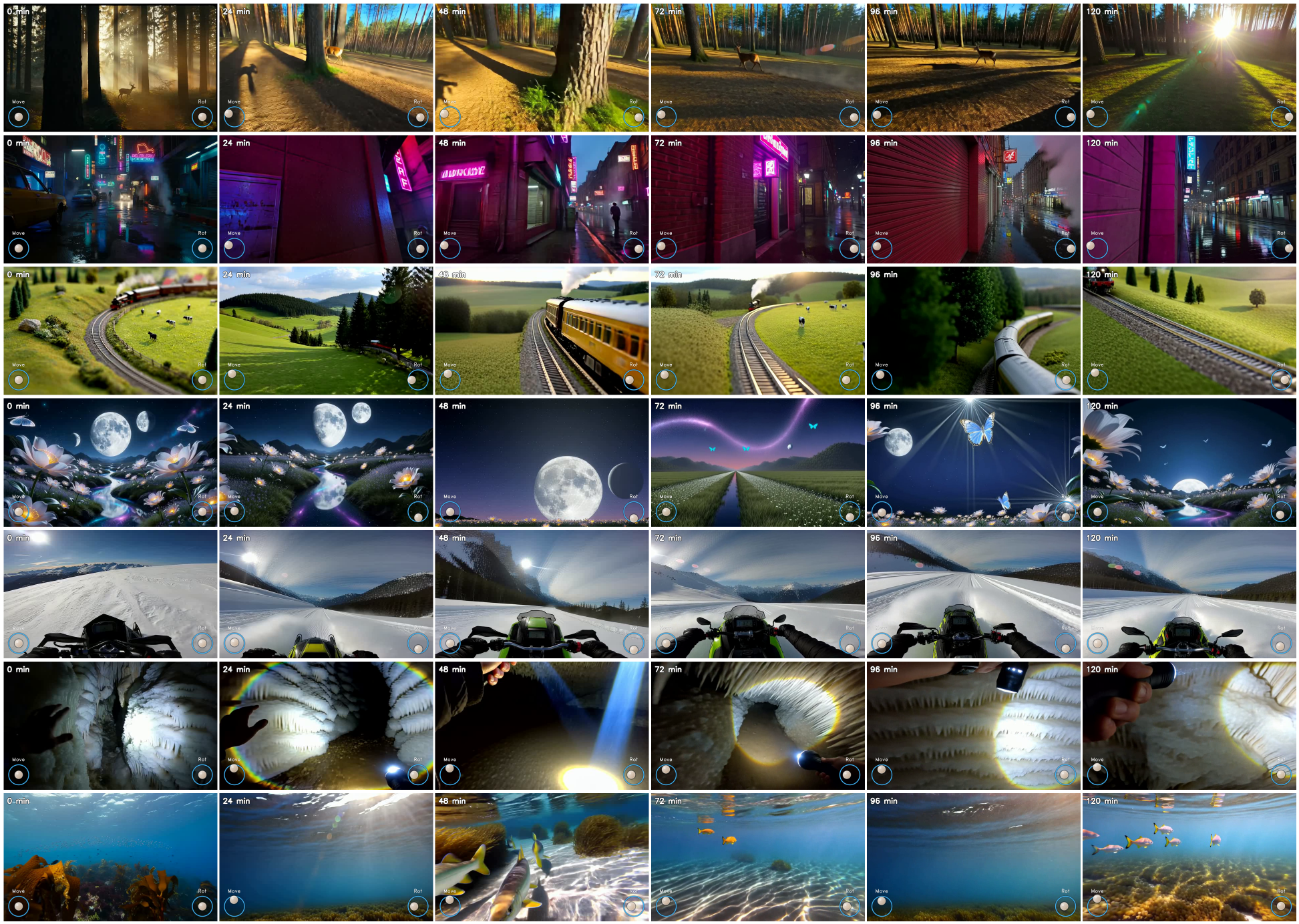}
  \vspace{-4mm}
  \caption{\textbf{Two hours of uninterrupted generation.} Representative two-hour rollouts under continuous camera control, generated in three steps per chunk without classifier-free guidance. Each row shows six frames sampled uniformly over the full session, illustrating sustained coherence over hour-scale generation.}
  \label{fig:l}
\vspace{-4mm}
\end{figure}

Interactive world models must jointly support persistent memory, responsive interaction, and long-horizon generation~\citep{genie,genie3,matrixgame2,worldplay,yume15}. A revisited scene should remain consistent with its earlier appearance; camera motions and text instructions should take effect promptly; and a session should extend from seconds to minutes or even hours. While substantial progress has been made on each capability in isolation, combining all three remains challenging. Their conjunction exposes a systems-level question that model scaling alone does not resolve~\citep{yin2025towards}: how to maintain persistent world state without an ever-growing context, and how to equip a low-latency generator with the capabilities required for long-horizon interaction.

Current interactive world models typically place two demanding requirements on the denoiser: preserving persistent world state and generating responses at interactive latency. When past observations are retained as additional context frames or an accumulated key-value cache~\citep{diffusionforcing,framepack,hunyuangamecraft,longlive}, the cost of each denoising step grows with the session history. Windowing and cache eviction can bound this cost only by discarding information~\citep{selfforcingpp}, while retrieval and streaming conditioning still operate under a finite denoiser-side context budget~\citep{relic,matrixgame3,streamdiffusionv2}. Interactive generation introduces a different constraint: low latency requires inference in only a few denoising steps, which is typically achieved by distillation from a slower teacher~\citep{dmd,dmd2,causvid,selfforcing}. The resulting student is therefore constrained by the temporal horizon and conditioning capabilities represented in its teacher supervision, limiting both long-term consistency and responsiveness to changes introduced during a session. These limitations suggest that persistent state and interactive capabilities need not both be carried by the deployed denoiser. \textbf{Alaya-EVOKE (Evoke) decouples them: persistent spatial state is maintained in an external world state bank, while the teacher is explicitly designed to provide the long-horizon and dynamically conditioned supervision required by the few-step student.} This separation forms the central design principle of Evoke and enables bounded-cost, long-running interaction.

For persistent world state, Evoke uses an explicit, bounded world state bank indexed by camera pose. Rather than retaining the observed history within the denoiser through context frames, key-value caches, or learned retrieval, the store records scene geometry externally and queries it directly from the current viewpoint. Previously observed surfaces that re-enter the camera frustum are rendered into the current view and provided to the generator as pixel-space conditioning, extending warp-based history conditioning~\citep{warpashistory} from short-range frame correspondences to persistent geometric memory. As a result, the denoiser operates with a fixed number of history and rendered memory frames, keeping its token budget independent of session duration. The world state bank is itself bounded and supports explicit read, write, and eviction; the current implementation retains a finite temporal window of geometry, providing constant-cost recall within the retained coverage rather than indefinite memory of every previously visited location.

The world state bank follows a geometry-based construction similar to prior work~\citep{warpashistory,gen3c,lyra2,spatialmemory}: generated observations are lifted into scene geometry and later rendered from the current camera pose to provide view-aligned conditioning. This explicit geometric representation keeps persistent state outside the denoiser while preserving direct access to previously observed regions.

A distilled student can acquire only capabilities expressed by its teacher supervision. Yet distillation is commonly treated primarily as an acceleration mechanism~\citep{dmd,dmd2,apt,causvid}, leaving key teacher properties largely fixed: attention may remain bidirectional and quadratic, training clips short, and text conditioning constant throughout a sequence. Recent work has examined the mismatch between bidirectional teachers and causal students~\citep{causalforcing}.
Evoke focuses on two complementary teacher-side design variables: supervision horizon and conditioning schedule. Short-horizon supervision provides no joint constraint over distant moments, while a single global prompt does not expose how newly introduced instructions should affect an ongoing rollout. The supervision horizon is particularly important because different failure modes emerge at different temporal scales. \emph{Degradation drift}, such as exposure growth, saturation shifts, and texture degradation, alters local statistics and can therefore be detected within relatively short windows. Existing long-video objectives can substantially improve such local stability~\citep{selfforcingpp,rollingforcing}, but this improvement may coincide with reduced temporal dynamics or diversity, where suppressing variation also makes short-window statistics easier to preserve.
\emph{Content drift} presents a different challenge: scene identity, object appearance, or spatial layout may evolve gradually while every local window remains individually plausible. Such inconsistencies become apparent only when sufficiently distant moments are compared and therefore cannot be resolved simply by enforcing stronger local stability. This distinction makes supervision horizon an explicit teacher-side design variable: extending the teacher horizon exposes long-range inconsistencies while retaining the dynamics required for an evolving world. Per-chunk conditioning complements long-horizon supervision by exposing the teacher to prompt changes and newly introduced events within the same rollout. Evoke therefore designs the teacher jointly for long-range consistency and dynamic conditioning, providing both capabilities for subsequent transfer to the few-step student. Whether the benefit of increasing supervision horizon eventually saturates is evaluated empirically in Section~\ref{sec:exp-session}.

\begin{figure}[tb]
  \centering
  \includegraphics[width=\linewidth]{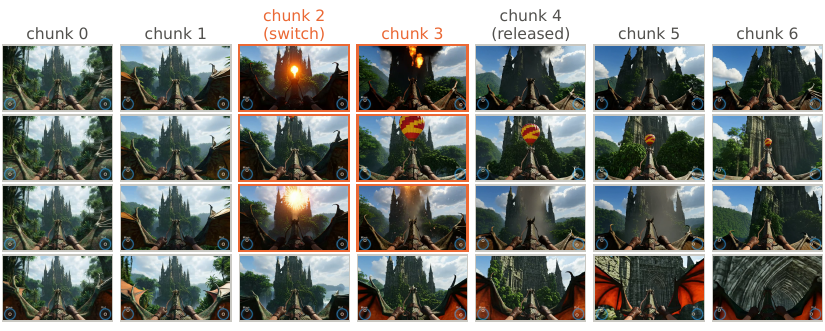}
  \vspace{-5mm}
\caption{\textbf{Timed prompt switching modifies the sky while preserving anchored scene structure.} Using the same input image, camera trajectory, and three-step CFG-free student, rows differ only in text. Rows~1 to 3 introduce a fireball, hot-air balloon, and fireworks, respectively, at chunk~$2$ and remove them at chunk~$4$ (amber borders denote active chunks); the objects appear and disappear while the castle, jungle, and dragon remain consistent. Row~4 uses a static castle prompt with no evocation.}
  \label{fig:j2}
\vspace{-4mm}
\end{figure}

These requirements lead to a redesigned Evoke Teacher for long-horizon supervision. The teacher adopts chunk-wise sparse attention~\citep{moga,mixtureofcontexts}, where each chunk accesses bounded local context, a small set of distant frames, and a linear-attention global state~\citep{gla,kimilinear,sanavideo}, reducing attention growth from quadratic to linear with sequence length. The same chunk structure assigns an independent text condition to each chunk, allowing instruction changes to be represented within a single rollout. A 30-second full-window distribution-matching objective~\citep{dmd,dmd2}, applied to self-forced rollouts~\citep{selfforcing}, jointly transfers long-range supervision and dynamic conditioning to a three-step, CFG-free~\citep{cfg} student. At inference time, the resulting student supports scheduled prompt changes during an ongoing session, enabling elements to be introduced or withdrawn on demand, a capability we refer to as \emph{evocation} (Fig.~\ref{fig:j2}).

%
%

Open-ended generation additionally requires the recurrent process itself to remain independent of elapsed time. A rollout represented on a continuously growing positional axis~\citep{rope} may eventually exceed the positional range encountered during training~\citep{selfforcingpp}.
Evoke instead reuses the same local positional layout for every generated chunk, while persistent world state is accessed through camera pose rather than temporal position. Session duration therefore increases only the number of recurrent generation steps, without expanding the positional range or denoiser context of any individual call.

The three-step student generates a $1.5$\,s chunk in $2.11$\,s on a single H200 at $384\times640$, with the world state bank enabled. Figure~\ref{fig:l} shows seven uninterrupted two-hour rollouts under continuous camera control, demonstrating coherent hour-scale generation. Evoke further achieves state-of-the-art performance on WBench~\citep{wbench} while remaining competitive with many-step systems on VBench-2.0~\citep{vbench2} and VBench-Long~\citep{vbench,vbenchpp}, despite using only three sampling steps and no classifier-free guidance. Quantitative long-session evaluation is reported separately in Sec.~\ref{sec:exp-session}, together with ablations on supervision horizon and conditioning granularity. These experiments isolate the two teacher-side design variables, measuring how long-horizon supervision affects content drift and how per-chunk conditioning governs mid-session responsiveness. Our contributions can be summarized as follows.

\begin{list}{}{%
\setlength{\leftmargin}{0pt}\setlength{\rightmargin}{0pt}%
\setlength{\labelwidth}{0pt}\setlength{\labelsep}{0pt}%
\setlength{\itemindent}{0pt}\setlength{\listparindent}{0pt}%
\setlength{\topsep}{4pt}\setlength{\itemsep}{3pt}\setlength{\parsep}{0pt}}
\item (i)~We formulate interactive world generation as a bounded recurrent process that decouples persistent world state from the denoiser: scene geometry is maintained in an external, camera-indexed world state bank, keeping the denoiser context and positional range independent of session duration.

\item (ii)~We redesign the teacher for long-horizon interactive supervision through chunk-wise sparse attention and per-chunk conditioning, enabling efficient supervision over long temporal horizons while exposing both long-range content drift and mid-sequence instruction changes.

\item (iii)~We transfer these capabilities to a three-step, CFG-free student using a $30$-second long-horizon distribution-matching objective under self-forced rollouts, yielding bounded-cost, open-ended generation with strong long-horizon consistency and responsive mid-session control.
\end{list}

\section{Related work}
\label{sec:related}

Interactive video world models~\citep{genie,genie3,gamengen,matrixgame,cosmos,navigationwm,oasis3} differ fundamentally in how they represent and retain information across a running session. One line of work maintains history within the denoiser, using additional context frames or an expanding key-value cache~\citep{diffusionforcing,framepack,hunyuangamecraft}, and controls the resulting growth in per-step cost through windowing, cache eviction, retrieval, or streaming video-to-video conditioning~\citep{selfforcingpp,rollingforcing,relic,matrixgame3,streamdiffusionv2}; several systems have demonstrated substantially extended generation horizons~\citep{genie3,matrixgame2,worldplay,longlive,skyreelsv2,ttt,svi}.
A second line externalizes part of this history through geometry~\citep{warpashistory,gen3c,lyra2,spatialmemory}: previously generated observations are lifted into a scene representation~\citep{nerf,gaussiansplatting}, using monocular depth or feed-forward geometry estimation~\citep{depthanything3,vggt}, and rendered into the target view, allowing revisited content to re-enter the generator as view-aligned pixel conditioning rather than as an ever-growing token history.
A third line targets interactive inference through few-step distillation, building on few-step samplers developed for image diffusion~\citep{progressivedistillation,lcm,consistencymodels}. Distribution-matching objectives~\citep{dmd,dmd2,causvid} and self-forcing~\citep{selfforcing} compress slower teachers into low-step students, alongside adversarial post-training~\citep{apt}, while recent work further examines the mismatch between bidirectional teacher supervision and causal student inference~\citep{causalforcing}.

Evoke builds on the latter two directions while shifting the design focus from individual mechanisms to where persistent state and transferable capabilities reside. Following geometry-based conditioning~\citep{warpashistory,gen3c}, Evoke renders previously observed scene content into the current view; however, the underlying world state is maintained in an explicit external world state bank with bounded read, write, and eviction, keeping the denoiser context independent of session duration.
On the distillation side, Evoke adopts distribution matching~\citep{dmd,selfforcing} but treats the teacher itself as a design variable rather than merely a source of high-quality targets. Beyond attention causality~\citep{causalforcing}, the teacher is explicitly structured around two properties required for long-horizon interaction: supervision horizon and conditioning schedule. Long-horizon supervision exposes content inconsistencies that remain locally plausible over short windows, while per-chunk conditioning allows instruction changes to be represented within the same rollout. These properties are then transferred to the few-step student through long-horizon distribution matching.

\section{Evoke}
\label{sec:method}

\begin{figure}[t]
  \centering
  \includegraphics[width=\linewidth]{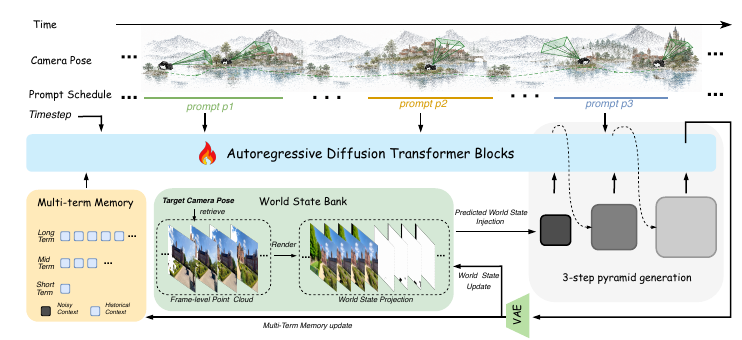}
  \vspace{-5mm}
  \caption{\textbf{With bounded retention, per-step cost does not grow with session
  length.} The camera pose reads the world state bank into the target view; the student
  injects that render at the coarsest of its three coarse-to-fine evaluations, alongside a
  short parametric history and the current text condition. The emitted chunk then updates
  the store and the history.}
  \label{fig:n}
\vspace{-4mm}
\end{figure}

Evoke realizes long-horizon interactive generation through two complementary components: a bounded recurrent student with externalized world state, and a teacher designed for long-horizon interactive supervision. At each recurrent step, the three-step student generates one video chunk from a short local history, view-aligned geometry retrieved from an external camera-indexed world state bank, and a per-chunk text condition, without increasing the context length or positional span of any individual generation step as the session grows (Fig.~\ref{fig:n}). The teacher combines chunk-wise sparse attention with per-chunk conditioning to provide supervision across distant moments and changing instructions, and these capabilities are transferred to the student through a $30$-second long-horizon distribution-matching objective applied under self-forced rollouts. The following sections formalize the recurrent formulation, motivate the required supervision horizon, and detail the teacher, world state bank, distillation procedure, and inference process.
%
%
%

\subsection{Recurrent Session Formulation}
\label{sec:setup}
%
\begin{figure}[t]
  \centering
  \includegraphics[width=\linewidth]{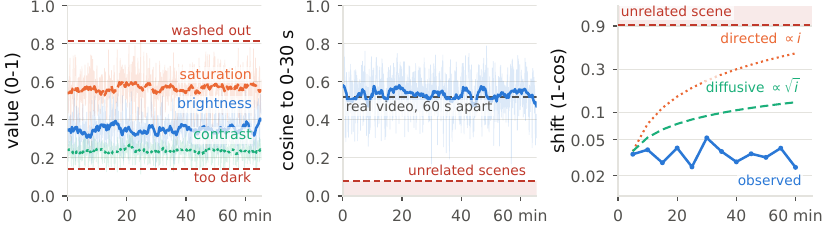}
  \vspace{-5mm}
  \caption{\textbf{An hour-long session stays bounded rather than degrading.} One continuous
  $65.5$\,min Evoke rollout, three steps per chunk without classifier-free guidance
  ($2619$ chunks; faint per-chunk, bold two-minute mean). \textbf{(a)} Color is flat over the hour. \textbf{(b)} Scene identity plateaus at
  cosine $0.523$, the level real footage scores against itself $60$\,s apart.
  \textbf{(c)} Opening-window shift stays far below unrelated-scene drift. A stability
  claim, not fidelity ($n=1$).}
  \label{fig:a}
\vspace{-4mm}
\end{figure}

Evoke represents long-running generation as a bounded recurrent process over video chunks. At recurrent step $k$, the student generates a chunk $x_k$ containing $F=9$ latent frames, corresponding to $36$ pixel frames or $1.5$\,s at $24$\,fps. Each step is conditioned on a camera trajectory $\mathcal{P}_k=\{P_t\}$~\citep{cameractrl,cameractrl2}, containing camera-to-world extrinsics and intrinsics resampled to $24$\,fps, and a text condition $c_k$ that may vary across steps. A session $x_{1:K}$ is therefore formed by repeatedly applying the same fixed-shape generation interface for an arbitrary number of steps.

The recurrent interface separates transient denoiser context from persistent geometric world state, the loop drawn in Fig.~\ref{fig:n}. Let $h_k$ denote a bounded local history and $M_k$ the world state bank. At each step, geometry relevant to the current camera trajectory is first rendered from $M_k$, the next video chunk is then generated from the rendered geometry, local history, and text condition, and the resulting observation is written back to the store:
\begin{equation}
r_k = \operatorname{Read}(M_k,\mathcal{P}_k), \qquad
x_k \sim p_\theta(\cdot \mid r_k,h_k,c_k), \qquad
M_{k+1} = \operatorname{Write}(M_k,x_k,\mathcal{P}_k).
\label{eq:session}
\end{equation}
Both the local history and world state bank operate under fixed budgets. Consequently, extending a session increases only the number of recurrent calls, without increasing the context length, positional span, or computational footprint of an individual call. Figure~\ref{fig:a} illustrates an hour-scale session under this formulation, and Sec.~\ref{sec:exp-session} evaluates the resulting runtime behavior.

This bounded recurrence determines how Evoke can continue generating for long durations, but does not by itself ensure long-range content consistency or responsiveness to newly introduced instructions. These capabilities arise from two complementary sources: the supervision used to train $p_\theta$, which determines what long-horizon behavior the student can acquire, and the external state $M_k$, which preserves previously observed scene information across recurrent steps. The following sections analyze these two roles separately, beginning with the supervision horizon.

\subsection{Long-Horizon Supervision}
\label{sec:horizon}

The bounded recurrent formulation keeps each inference step independent of session duration, but it does not by itself prevent errors from accumulating across recurrent steps. The remaining question is how much of a rollout must be visible to the training objective in order to constrain long-range behavior. Let $q_\theta$ denote the distribution of student trajectories $x_{1:K}$ and $p$ the corresponding data distribution. Under self-forced distribution matching~\citep{dmd,dmd2,selfforcing}, supervision is evaluated over windows of $W$ consecutive chunks:
\begin{equation}
\mathcal{L}_{W}(\theta)
=
\mathbb{E}_{k}
\left[
D\!\left(
q_{\theta}^{(k:k+W-1)}
\,\middle\|\,
p^{(k:k+W-1)}
\right)
\right],
\label{eq:window-objective}
\end{equation}
where $D$ denotes the divergence whose gradient is estimated from the teacher and critic scores. By construction, this objective depends only on trajectory statistics observable within a $W$-chunk window. Two trajectories may therefore receive identical supervision at this horizon even if their behavior diverges over longer temporal spans. The supervision horizon $W$ consequently determines not only how much generated history is exposed during training, but also which forms of rollout drift can produce a learning signal.

This distinction is important because long-rollout errors arise at different temporal scales. \emph{Degradation drift}, such as exposure shifts, saturation changes, and progressive texture degradation, eventually becomes visible within a local window. Its cause, however, may precede that window. During a self-forced rollout, the student conditions on its own generated history rather than ground-truth observations~\citep{selfforcing,selfforcingpp}, so earlier errors perturb the conditioning distribution encountered by subsequent denoising steps. A short window may capture the resulting degradation while exposing only a narrow range of the preceding perturbations that produced it. Increasing $W$ therefore broadens the distribution of rollout-induced histories under which the student is supervised and improves its ability to remain stable under accumulated conditioning shift.
\emph{Content drift} imposes an additional constraint. Scene identity, object appearance, or spatial layout may evolve gradually while every short window remains individually plausible. Such errors become identifiable only when sufficiently distant moments are jointly covered by the supervision window~\citep{selfforcingpp,rollingforcing}.
Long-horizon supervision thus serves two complementary roles: it enlarges the range of rollout perturbations under which the recurrent generator learns to remain stable, and it exposes long-range inconsistencies that cannot be identified from local trajectory statistics alone.

Not all long-range inconsistency should be resolved by extending the supervision horizon. For \emph{temporal content drift}, the correct evolution is governed by dependencies in the data distribution across distant moments; no particular past observation uniquely specifies the desired continuation. This failure is therefore naturally addressed through longer-horizon supervision. \emph{Spatial revisit inconsistency} has a different source. When the camera returns to a previously observed surface, the relevant state has already been produced by the session, and the problem is to make that specific observation available again. Evoke accordingly assigns these two cases to different mechanisms: long-horizon teacher supervision constrains temporal evolution, while the world state bank preserves observed scene state and retrieves it through camera pose. A longer supervision window cannot substitute for an unavailable observation, and stored geometry alone cannot determine how unobserved content should evolve.

Importantly, the required supervision horizon need not scale with the final session duration. The goal is not to train the student to recover from arbitrarily corrupted histories, but to make its recurrent dynamics stable over the range of perturbations encountered during normal generation. As the student becomes more robust to accumulated conditioning shift, its own rollouts remain within a progressively better behaved region of trajectory space, reducing the need to cover increasingly extreme failure states. Extending $W$ should therefore provide large gains while it continues to expose new rollout perturbations and previously invisible long-range inconsistencies, followed by diminishing returns once the relevant temporal scale is covered. Evoke uses approximately $30$\,s of long-horizon supervision and tests this prediction by sweeping $W$: resistance to drift should improve with the supervision horizon and then approach saturation rather than scale proportionally with the desired rollout length. 

This analysis determines the requirements placed on the remaining components. The teacher must support sufficiently long windows at a cost compatible with repeated distillation queries, motivating the efficient long-sequence architecture in Sec.~\ref{sec:teacher}. Its supervision must also expose the time-varying conditioning encountered at deployment, since a teacher evaluated under a single global prompt cannot demonstrate how an instruction introduced mid-session should affect the rollout. Finally, previously observed scene geometry must remain explicitly accessible when revisited, which motivates the external world-state mechanism in Sec.~\ref{sec:spatial}.

\subsection{Long-Horizon Interactive Teacher}
\label{sec:teacher}

%
%
\begin{figure}[t]
  \centering
  \includegraphics[width=\linewidth]{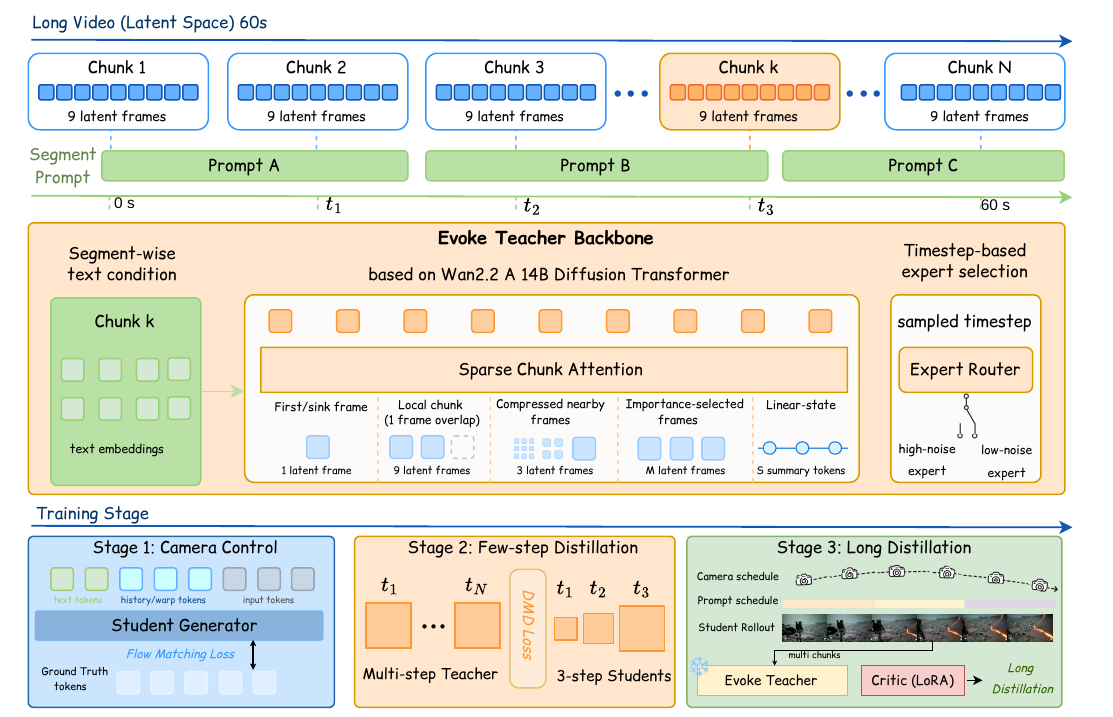}
  \vspace{-5mm}
  \caption{\textbf{Chunk-wise sparse attention makes long, time-varying supervision
  affordable.} Each query chunk of a $60$\,s clip attends to a fixed set of key sources
  and reads the text segment covering it, so attention cost grows linearly in length; the
  sampled timestep selects the high- or low-noise expert. Below: camera-control training
  and few-step distillation precede long-rollout distribution matching against the shared
  teacher/critic backbone.}
  \label{fig:m}
\vspace{-4mm}
\end{figure}

%
%
\begin{figure}[t]
  \centering
  \includegraphics[width=\linewidth]{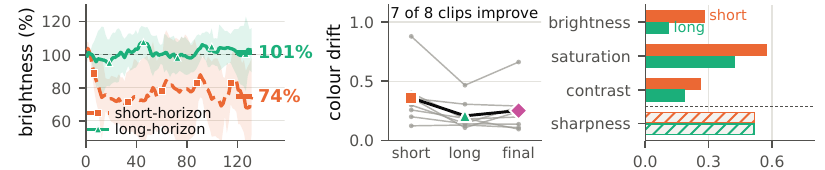}
  \vspace{-5mm}
  \caption{\textbf{A student inherits photometric stability from a long-horizon teacher.} Two
  three-step students with identical distillation recipes and teachers differing only in
  temporal horizon; clips, trajectory, prompt, seed and inference flags are shared, so only
  the checkpoint differs. \textbf{(a)} The short-horizon student settles at $74\%$ of its
  opening brightness, the long-horizon one at $101\%$. \textbf{(b)} Seven of eight clips
  improve (Wilcoxon $p=0.016$). \textbf{(c)} Brightness separates the two most; sharpness
  not at all. Claim: exposure stability, not image quality.}
  \label{fig:f2}
\vspace{-4mm}
\end{figure}

The previous section shows that the long-horizon behavior acquired by a few-step student is directly limited by the supervision expressed by its teacher. Evoke therefore treats the teacher not as a fixed source of high-quality scores, but as a model whose supervision is explicitly designed for long-horizon interactive generation. Figure~\ref{fig:f2} provides a controlled comparison: two students are distilled with identical recipes from teachers that differ only in temporal horizon, and the student supervised by the long-horizon teacher exhibits substantially greater stability over extended rollouts; Sec.~\ref{sec:exp-teacher} provides the full analysis. The Evoke Teacher is built on the $14$B Wan2.2 A14B diffusion transformer~\citep{dit,wan,wan22} and retains its high-noise and low-noise experts selected according to the sampled timestep.
The teacher and critic share the same backbone rather than residing as two separate models. Enabling a LoRA adapter~\citep{lora} yields the critic, while disabling it yields the teacher, so the teacher and critic scores are always computed within the same expert.

To make long-window supervision computationally practical, the Evoke Teacher partitions a sequence into chunks of nine latent frames and applies chunk-wise sparse attention~\citep{moga,mixtureofcontexts}. Each query chunk accesses a bounded set of sources: a first-frame global sink, local context with a one-frame overlap, spatially compressed nearby frames, a small set of selected distant frames, and a global state accumulated through linear attention~\citep{gla,kimilinear,sanavideo}, as illustrated in Fig.~\ref{fig:m}. Because the amount of information accessed by each chunk remains bounded, the resulting attention computation grows approximately linearly rather than quadratically with sequence length, as in other sparse-attention designs for long video generation~\citep{radialattention}, making repeated long-window scoring feasible during distillation. The same chunk partition provides an independent text-conditioning context for every chunk, allowing prompt changes to be represented within a single long sequence. This factorization also brings the teacher closer to the student's chunk-wise recurrent inference, since local context and retrieved distant frames are drawn from the past while only limited non-causal paths remain. The mismatch between bidirectional teacher attention and causal student inference has been studied previously~\citep{causalforcing}; Evoke instead focuses on the complementary axes of supervision horizon and conditioning schedule.

Evoke transfers this long-horizon supervision to the few-step student through full-window distribution matching distillation (DMD)~\citep{dmd,dmd2}. Training starts from one ground-truth prefix chunk $x_0^{\mathrm{gt}}$ followed by a self-forced rollout of $20$ generated chunks $x_1,\ldots,x_{20}$, yielding $21\times9=189$ latent frames, or $753$ pixel frames corresponding to approximately $31.4$\,s. Since the student performs three function evaluations per chunk, constructing the rollout requires $60$ student forward passes. Rather than sampling a short window from this trajectory, the teacher and critic jointly score all $189$ latent frames, so the supervision horizon spans the complete self-forced rollout. Let $\hat{x}_0$ denote the student prediction, $s_{\mathrm{real}}$ the teacher prediction, and $s_{\mathrm{fake}}$ the critic prediction. We define the score difference $\Delta s$ and normalize it by $\nu$ over the region $\Omega$ that receives the distribution-matching gradient:
\begin{equation}
\Delta s = s_{\mathrm{fake}} - s_{\mathrm{real}},
\qquad
\nu =
\operatorname{mean}_{\Omega}
\left[
\left|
\hat{x}_0 - s_{\mathrm{real}}
\right|
\right],
\qquad
\mathcal{L}_{\mathrm{gen}}
=
\frac{1}{2}
\left\|
\hat{x}_0 -
\left(
\hat{x}_0 - \frac{\Delta s}{\nu}
\right)^{\mathrm{detach}}
\right\|_2^2 .
\label{eq:dmd}
\end{equation}
The mask $\Omega$ excludes the ground-truth prefix and the first generated chunk $x_1$. At this boundary, the teacher treats its first chunk according to the image-model first-frame distribution, whereas the student performs video continuation; directly matching the two introduces boundary flicker. The DMD gradient is therefore applied only to the remaining generated chunks $x_2,\ldots,x_{20}$, and $\nu$ is computed over the same mask so that excluded frames do not affect the update scale.

An important distinction is that the \emph{supervision horizon} need not equal the \emph{gradient horizon}. History is detached between consecutive rollout chunks, restricting each backward graph to a single chunk and allowing chunk-level gradients to be computed independently. This shortens the gradient horizon without shortening the supervision horizon: the teacher and critic still evaluate the complete $31.4$\,s trajectory jointly, so each local student update is derived from a distribution discrepancy defined over the full rollout. Long-horizon supervision therefore does not require backpropagation through the entire trajectory. Detachment bounds activation memory, while full-window scoring preserves the long-range training signal.

The teacher must also express the time-varying conditioning encountered at deployment. At inference, the student models
\begin{equation}
p_\theta
\left(
x_k
\mid
x_{<k},
\mathcal{P}_k,
c_k
\right),
\end{equation}
where the text condition $c_k$ may change at every recurrent step. A teacher evaluated under a single global prompt instead supervises a fixed-condition trajectory distribution and never demonstrates how a newly introduced instruction should affect an ongoing rollout. Evoke therefore assigns text conditions at the chunk level and includes prompt transitions within the training sequence, aligning both the supervision horizon and conditioning schedule with deployment. Training captions are segmented at $12$\,s intervals and mapped to their corresponding latent chunks, while each student rollout chunk receives the text condition associated with its temporal position. At inference, the same representation is exposed as a timed prompt schedule, allowing text-driven elements to be introduced or withdrawn during an ongoing session, a capability referred to as \emph{evocation}. Figure~\ref{fig:j2} shows a controlled example in which only the timed text schedule changes while the camera trajectory, warp history, and random seed remain fixed, and Figure~\ref{fig:o} in the appendix shows the same interface at the teacher's own horizon: four-minute teacher rollouts driven by twelve consecutive instructions of $20$\,s each. This formulation also enables a direct conditioning ablation: with the supervision horizon $W$ held fixed, per-chunk teacher conditioning can be replaced by a single global prompt to isolate its effect on mid-session responsiveness, as evaluated in Sec.~\ref{sec:exp-cond}. 

Pure distribution matching does not explicitly enforce adherence to the requested camera trajectory $\mathcal{P}_k$. Evoke therefore retains the supervised warp-conditioning objective from the preceding training stage as an auxiliary regularizer for camera controllability, while long-horizon distribution matching transfers the teacher's long-range stability and dynamic-conditioning capabilities. Full-window training further uses sequence parallelism~\citep{ulysses}, activation recomputation, and independent chunk-level backward graphs to control the cost of the $189$-frame teacher scoring and $20$-chunk student rollout; the corresponding systems implementation and efficiency analysis are reported in the experiments.

\subsection{Geometric World State}
\label{sec:spatial}

Evoke externalizes persistent world state that cannot remain within the bounded denoiser context~\citep{spatialmemory}. The student retains only the most recent $19$ latent frames, corresponding to approximately $3.2$\,s, organized into long-, mid-, and short-range history tiers of $16$, $2$, and $1$ frames, respectively. Observations outside this local history are no longer represented in the denoiser context, even though they may become relevant again when the camera revisits a previously observed region. Other designs keep such recalled history inside the denoiser, through a compressed cache or retrieved memory tokens~\citep{relic,matrixgame3}. Evoke therefore maintains an external world state bank $M_k$, using camera pose both as a control signal for generation and as an address for retrieving previously observed scene content. This realizes the spatial component identified in Sec.~\ref{sec:horizon}: long-range temporal evolution is learned through supervision, while previously observed spatial state is recovered explicitly.

At each recurrent step, newly generated observations are written into $M_k$ as world-space geometry. A monocular depth model~\citep{depthanything3} estimates depth for $12$ frames of the generated chunk under the known camera trajectory.
The resulting depth maps are unprojected with the corresponding camera intrinsics and extrinsics and appended to the world state bank. Geometry from different chunks is inserted independently, which avoids the scale drift and rendering artifacts observed when depth estimates are fused across long generated sequences. To read the store, the current camera pose directly determines which stored observations are geometrically relevant. Stored source views are ranked by co-visibility with the target view, and up to eight sufficiently distinct sources are selected and rendered through batched projection with $z$-buffering, following prior warp-conditioned generators~\citep{warpashistory,gen3c}. The read operation returns both a view-aligned warped observation and a per-pixel visibility mask. In contrast to learned retrieval, recall is determined directly by camera geometry, which is appropriate for spatial revisit because the desired content has already been observed rather than inferred anew.

%
\begin{figure}[tb]
  \centering
  \includegraphics[width=\linewidth]{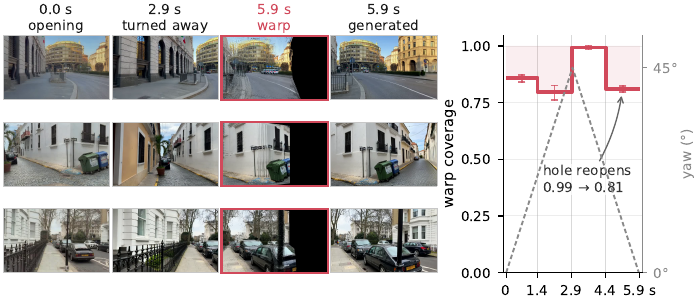}
  \vspace{-5mm}
  \caption{\textbf{The world state bank puts a place back when the camera returns.} Three held-out
  street scenes follow one authored $5.9$\,s trajectory, generated by Evoke in three steps per
  chunk without classifier-free guidance: a $45^{\circ}$ turn that carries
  the opening view out of frame, and back. \emph{Left to right:} opening view; turn
  endpoint; the bank's return render; the frame generated from it---the black band is what
  the bank no longer holds and the model must inpaint. \emph{Right:} retaining only the
  latest $1.5$\,s of geometry buys nothing while the view is new and costs $0.173$ of
  coverage at the return. Coverage bounds recall from above; it is not a fidelity measure.}
  \label{fig:k}
\vspace{-4mm}
\end{figure}

The visibility mask controls how the rendered state enters the denoiser. Regions supported by stored geometry retain informative warp conditioning, whereas unsupported regions are suppressed by increasing the warp noise level; pixels with visibility below $0.5$ are assigned $\sigma=1$ and therefore contribute no visual information. Supported regions receive substantially lighter noise, with $\sigma\in[0,0.135]$. The same visibility signal is pooled at the patch resolution of each history tier and used to remove unsupported history tokens from the denoiser sequence, a visible-token selection introduced by prior work~\citep{warpashistory}. Consequently, geometric memory contributes only where the store provides reliable coverage: observed surfaces can be propagated into the current view, while uncovered regions remain available for synthesis rather than being constrained by unreliable warps.

The external state is operated under a fixed retention budget in long-session inference. The configuration used for hour-scale experiments retains $2160$ pixel frames, corresponding to $90$\,s of geometry; because every third frame is ingested, the active source pool contains at most $720$ frames. Together with the fixed upper bound on retrieved source views, this keeps both storage and per-step geometric conditioning bounded with respect to session duration. The resulting guarantee is therefore persistent recall within retained geometric coverage rather than permanent recall of every location ever observed. When a retained surface is revisited, its state can be recovered through camera pose without extending the denoiser context, while observations outside the retention budget no longer constrain subsequent generation. Figure~\ref{fig:k} illustrates this behavior by rendering a previously observed region when the camera returns to its stored geometry.

\subsection{Bounded Three-Step Inference}
\label{sec:runtime}

Evoke generates each chunk with three CFG-free~\citep{cfg} denoising evaluations over a coarse-to-fine latent pyramid~\citep{pyramidflow} at resolutions $12\times20$, $24\times40$, and $48\times80$.
Geometric conditioning is injected only at the coarsest stage, where it establishes large-scale spatial structure before the higher-resolution stages refine appearance and detail. Visibility-based pruning further removes unsupported geometric tokens, so the additional conditioning cost depends on the coverage of the world state bank rather than on session duration.

Together with bounded local history, bounded geometric retrieval, and local positional indices, each recurrent call operates within a session-independent context and positional range. Extending a session therefore increases only the number of recurrent calls, not the size of an individual call. Runtime and geometric-conditioning overhead are evaluated in Sec.~\ref{sec:exp-efficiency}.

\section{Experiments}
\label{sec:exp}

We evaluate Evoke along four dimensions: overall world-model and video-generation quality, stability and efficiency over hour-scale sessions, the effect of long-horizon teacher supervision, and the interaction between timed text control and persistent geometric memory. Together, these experiments assess both the quality of the three-step student and the design choices that enable it to operate continuously. Qualitative rollouts across egocentric, cinematic, embodied, and event-driven settings are collected in Fig.~\ref{fig:p} in the appendix.

\subsection{Experimental Setup and Benchmark Performance}
\label{sec:exp-setup}

\textbf{Models and protocol.}
The Evoke student is built on Helios~\citep{helios} and progressively trained for camera-controllable generation and few-step inference, followed by long-horizon distillation (\emph{long-distill}) and a short post-distillation continuation (\emph{post-distill}). Training uses the Sekai video dataset~\citep{li2026sekai} together with additional internal video data. For controlled distillation experiments, we compare a few-step student distilled from a short-horizon teacher with its counterpart distilled using the Evoke long-horizon teacher; the released Evoke student is obtained by continuing the latter training pipeline. The Evoke Teacher is adapted from the $14$B Wan2.2 A14B diffusion transformer~\citep{wan,wan22} and trained for long-sequence supervision with the architecture described in Sec.~\ref{sec:teacher}. Unless stated otherwise, the released student uses three denoising evaluations per chunk without classifier-free guidance.

All runtime measurements use a single H200 at $384\times640$ resolution with the full VAE decoder and without inference-specific acceleration such as key-value caching, compilation, quantization, or a distilled decoder. Long-session quantitative evaluation uses eight $65.5$-minute rollouts of $2{,}619$ chunks each, with the world state bank bounded to $90$\,s of retained observations. Additional verification and detailed protocol variations are provided in the appendix.

%
%
%
%
\begin{table}[t]
\caption{WBench navigation split, $n=158$ cases, identical for all systems. All
scores $\in[0,100]$, higher is better; \textbf{bold} is best,
\underline{underline} second best. Group rows are the unweighted mean over their
group's metrics; peer numbers are as reported by their authors.}
\label{tab:wbench}
\centering
\renewcommand{\arraystretch}{1.15}
\setlength{\tabcolsep}{3pt}
\scriptsize
\resizebox{\ifdim\width>\linewidth\linewidth\else\width\fi}{!}{%
\begin{tabular}{@{}l *{9}{c} @{}}
\toprule
\textbf{Metric}
  & \textbf{Yume 1.5}
  & \textbf{Matrix-Game 2.0}
  & \textbf{\makecell{HY-World 1.5\\\textit{ar-distill}}}
  & \textbf{HY-GameCraft}
  & \textbf{\makecell{LingBot-World\\\textit{fast}}}
  & \textbf{\makecell{LingBot-World v2\\\textit{fast}}}
  & \textbf{Genie 3}
  & \textbf{Happy Oyster}
  & \textbf{Evoke (ours)} \\
\midrule
\multicolumn{10}{@{}l}{\textbf{Video Quality}} \\
  \quad Aesthetic   & 58.7 & 54.0 & 60.1 & 52.6 & 62.6 & \underline{64.4} & 51.6 & 56.6 & \textbf{66.12} \\
  \quad Imaging     & 63.3 & 60.3 & 65.4 & 58.7 & 63.8 & \underline{67.5} & 59.3 & 63.9 & \textbf{67.86} \\
  \quad Flickering  & 93.0 & \underline{94.6} & 93.5 & 93.7 & 92.4 & 91.4 & \textbf{95.0} & 94.0 & 94.31 \\
  \quad Dynamic     & \underline{96.8} & 94.9 & 91.1 & \underline{96.8} & 95.6 & 96.2 & 92.4 & 94.2 & \textbf{96.84} \\
  \quad Smoothness  & 97.0 & \textbf{98.2} & \underline{98.1} & 97.6 & 96.0 & 96.5 & 97.8 & 97.0 & 97.86 \\
  \quad HPSv3-Norm  & 57.0 & 41.0 & 60.5 & 38.3 & 65.7 & \textbf{74.6} & 55.2 & 58.3 & \underline{73.75} \\
  \quad \textit{Quality avg.\ (6)} & 77.63 & 73.83 & 78.12 & 72.95 & 79.35 & \underline{81.77} & 75.22 & 77.33 & \textbf{82.79} \\
\midrule
\multicolumn{10}{@{}l}{\textbf{Setting}} \\
  \quad Scene       & 53.1 & 49.4 & 53.5 & 50.6 & 63.4 & \underline{66.7} & 61.1 & 57.4 & \textbf{74.68} \\
  \quad Subject     & 91.7 & 84.9 & 90.8 & 82.5 & \underline{92.4} & 86.9 & 83.8 & 91.1 & \textbf{92.84} \\
  \quad \textit{Setting avg.\ (2)} & 72.40 & 67.15 & 72.15 & 66.55 & \underline{77.90} & 76.80 & 72.45 & 74.25 & \textbf{83.76} \\
\midrule
\multicolumn{10}{@{}l}{\textbf{Interaction}} \\
  \quad Navigation  & 72.0 & 80.6 & \textbf{87.5} & 67.8 & 79.4 & 82.8 & 73.3 & \underline{85.1} & 78.63 \\
\midrule
\multicolumn{10}{@{}l}{\textbf{Consistency}} \\
  \quad Background  & 90.3 & 86.9 & \textbf{92.7} & 86.5 & 90.9 & \underline{92.5} & 90.7 & 91.4 & 92.27 \\
  \quad Spatial     & 71.5 & 64.5 & \textbf{90.6} & 60.5 & 77.2 & 82.3 & 79.9 & 77.7 & \underline{84.26} \\
  \quad Gated Spatial & 71.4 & 64.5 & \textbf{84.9} & 60.5 & 76.9 & 78.7 & 78.4 & 75.8 & \underline{82.45} \\
  \quad Segment     & \underline{99.4} & 21.0 & \textbf{100.0} & \underline{99.4} & 98.1 & 98.1 & 93.6 & 96.2 & \textbf{100.00} \\
  \quad Perspective & 48.0 & 29.2 & 62.5 & 17.9 & \underline{82.8} & \textbf{84.5} & 54.5 & 75.0 & 69.74 \\
  \quad Subject     & 88.8 & 87.2 & 89.1 & 82.6 & 88.6 & 88.9 & 90.4 & \textbf{91.5} & \underline{91.03} \\
  \quad Geometric   & 88.0 & 86.1 & \underline{92.0} & 88.3 & 85.4 & 87.1 & 88.6 & 87.2 & \textbf{92.68} \\
  \quad Photometric & 83.3 & 81.3 & 83.1 & \textbf{85.0} & 79.1 & 79.8 & \underline{84.5} & 79.8 & 82.53 \\
  \quad \textit{Consistency avg.\ (8)} & 80.09 & 65.09 & \underline{86.86} & 72.59 & 84.88 & 86.49 & 82.58 & 84.33 & \textbf{86.87} \\
\midrule
\multicolumn{10}{@{}l}{\textbf{Physical}} \\
  \quad Causal Fidelity     & 72.7 & 59.3 & 74.0 & 68.3 & 72.5 & \underline{76.7} & 71.7 & 69.3 & \textbf{82.44} \\
  \quad Visual Plausibility & 57.7 & 55.0 & 58.6 & 56.5 & 58.8 & \underline{61.4} & 59.7 & 57.6 & \textbf{61.67} \\
  \quad \textit{Physical avg.\ (2)} & 65.20 & 57.15 & 66.30 & 62.40 & 65.65 & \underline{69.05} & 65.70 & 63.45 & \textbf{72.06} \\
\bottomrule
\end{tabular}
}

\end{table}

\textbf{Interactive world-model performance.}
On the $158$-case navigation split of WBench~\citep{wbench}, Evoke leads the Video Quality, Setting, and Physical group averages among the evaluated few-step systems~\citep{yume15,matrixgame2,hunyuanworld,hunyuangamecraft,lingbotworld,lingbotv2,genie3,happyoyster}, while remaining on par with the strongest result in Consistency. The largest gains occur in scene-level and causal-fidelity measures, consistent with the emphasis of Evoke on persistent scene state and long-running generation. Navigation and Perspective remain comparatively weaker, reflecting limitations of the current camera-control path that are analyzed separately.


\begin{table}[t]
\caption{Evoke on two public leaderboards, against the top-10 of each. Higher is better; the full tables are Tables~\ref{tab:vbench2}
and~\ref{tab:vbench-long} in the appendix.}
\label{tab:vbench}
\centering
\renewcommand{\arraystretch}{1.15}
\setlength{\tabcolsep}{6pt}
\small
\begin{tabular}{@{}l c c l l@{}}
\toprule
\textbf{Benchmark} & \textbf{Evoke} & \textbf{Rank} & \textbf{Leader} & \textbf{Nearest peer} \\
\midrule
VBench-2.0 & \textbf{66.77} & 1 of 10 & --- & Veo 3, 66.72 \\
VBench-Long & \textbf{85.11} & 7 of 10 & IPOW, 88.26 & Veo 3, 85.06 \\
\bottomrule
\end{tabular}

\vspace{2pt}
\begin{minipage}{\linewidth}\footnotesize
Evoke is sampled in $3$ steps with no classifier-free guidance, against peers
running their own many-step default samplers, so neither comparison is
step-matched. ``Leader'' is the best peer total and is ``---'' where Evoke
leads. Protocol deviations are stated in Appendix~\ref{app:vbench}.
\end{minipage}
\end{table}

\textbf{General video quality.}
Despite using only three CFG-free evaluations, Evoke remains competitive with many-step systems on both VBench-2.0~\citep{vbench2} and VBench-Long~\citep{vbench,vbenchpp}. It obtains an overall score of $66.77$ on VBench-2.0 and $85.11$ on VBench-Long, showing that the efficiency required for interactive generation does not come at the cost of a large degradation in general video quality. Full leaderboard results and per-dimension breakdowns are provided in Appendix~\ref{app:vbench}.

\subsection{Long-Session Stability and Bounded Runtime}
\label{sec:exp-session}
\label{sec:exp-longrun}

Figure~\ref{fig:l} qualitatively demonstrates uninterrupted two-hour generation under continuous camera control. For quantitative analysis, the eight $65.5$-minute sessions show that extending the rollout does not lead to progressively increasing visual degradation. Photometric statistics stabilize after an initial transient and exhibit little subsequent drift over the remaining session. Content descriptors change more rapidly at the beginning of a rollout and then evolve substantially more slowly; importantly, real-video controls exhibit comparable decorrelation over time. We therefore interpret these measurements as evidence against runaway long-session degradation rather than as evidence of permanent scene-identity preservation.

The computational behavior remains bounded over the same sessions. Once the fixed retention budget is filled, the active geometric source pool no longer grows with elapsed time, and the cost of a recurrent step remains stable throughout the rollout. Session duration therefore increases the number of recurrent calls without increasing the state or computation required by an individual call. The student denoises each $1.5$\,s chunk in $2.11$\,s on a single H200; this number measures diffusion wall clock, while the complete instrumented inference path, including geometry rendering and video I/O, is analyzed in Sec.~\ref{sec:exp-efficiency}.

\subsection{Effect of Long-Horizon Teacher Supervision}
\label{sec:exp-teacher}
\label{sec:exp-wsweep}

We next isolate the effect of the teacher by distilling students with otherwise matched recipes from short- and long-horizon teachers. Across paired long-rollout evaluations, the student supervised by the Evoke Teacher exhibits substantially stronger photometric stability than its short-teacher counterpart (Fig.~\ref{fig:f2}). This establishes that long-horizon teacher training can transfer improved resistance to progressive photometric drift into the few-step student. The measured content descriptor does not significantly separate the two students, so we restrict this conclusion to photometric stability rather than attributing all forms of long-range consistency to the teacher horizon.

We further examine whether this improvement can be explained simply by making drift detectable over a longer scoring window. Controlled scoring sweeps reveal little additional change in teacher-critic detectability once the window exceeds a short horizon, and no consistent sharp threshold emerges across the tested perturbations. The empirical benefit of the long-horizon teacher therefore does not reduce to a simple window-length detectability effect. Detailed sweeps, score-noise analysis, and additional teacher ablations are reported in the appendix. These results motivate treating long-horizon supervision as a property of the complete trained teacher and rollout distribution rather than attributing its effect to window length alone.

\subsection{Geometric Recall and Timed Interaction}
\label{sec:exp-memory}
\label{sec:exp-cond}
\label{sec:exp-efficiency}

\textbf{Pose-addressed recall.}
We first test whether the world state bank actually restores information that has left the denoiser context, measuring revisit PSNR between two $12$\,s windows rendered at identical camera poses. Across leave-and-return trajectories with different revisit intervals, recall remains near the far-pose floor measured within the same run when the retained geometry is shorter than the time spent away, and improves consistently once the retention window covers the revisit. Across the controlled conditions, $20$ of $21$ comparisons follow this predicted transition, and retention windows at least as long as the time away score $2.3$--$3.2$\,dB higher than shorter ones. The resulting $15.4$--$17.8$\,dB plateau indicates recognizable rather than pixel-faithful reconstruction, defining the scope of the current geometric memory. Figure~\ref{fig:k} visualizes the same mechanism at short timescale: when the camera returns to a previously observed view, retained geometry restores the corresponding regions instead of requiring them to be regenerated from the short denoiser history.

%
%
\begin{figure}[tb]
  \centering
  \includegraphics[width=\linewidth]{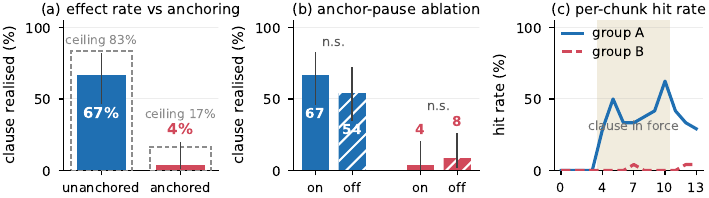}
  \vspace{-5mm}
  \caption{\textbf{A timed prompt switch introduces unanchored content; the anchored floor is not a
  switching property.} One checkpoint over $48$ schedules ($n=24$ per group). Blue requests
  a new object in unanchored space, red requests overwriting content pinned by the source
  frame and warp history. \textbf{(a)} Unanchored clauses are realized in $67\%$ of cases
  against an $83\%$ ceiling, anchored ones in $4\%$ against $17\%$---the anchored floor is
  mostly what this checkpoint renders at all, not the reach of switching. \textbf{(b)}
  Pausing the anchor separates neither group, but it also freezes the camera, so it does
  not isolate the anchor. \textbf{(c)} Per-chunk hit rate; shading marks assertion.}
  \label{fig:j}
\vspace{-4mm}
\end{figure}

\textbf{Timed text control.}
The per-chunk conditioning interface allows text instructions to change during an ongoing session. In the controlled evocation evaluation, a clause introduced mid-session is realized in $67\%$ of cases when it targets previously unanchored content, but only $4\%$ when realizing it would require replacing geometry already supported by the world state bank (Fig.~\ref{fig:j}). The qualitative examples in Fig.~\ref{fig:j2} show the same behavior: timed instructions can introduce and withdraw new elements while anchored scene structure remains stable. This contrast exposes a useful interaction between the two control paths: text governs content that remains free to evolve, whereas persistent geometry resists overwriting observations that have already been anchored. A matched short- versus long-teacher comparison does not significantly separate mid-session realization rates; accordingly, per-chunk conditioning is interpreted as the interface that expresses timed control rather than as an isolated causal explanation for the capability.

\textbf{Cost of geometric memory.}
The geometric path remains bounded by construction and adapts to the amount of retrieved coverage. Only a fixed number of source observations are rendered, and unsupported warp tokens are removed before denoising. Consequently, additional geometric-conditioning cost depends on how much previously observed content is visible in the current view rather than on how long the session has been running. Detailed wall-clock decomposition, the dependence on geometric coverage, camera-trajectory controls for revisit evaluation, and training-system efficiency are provided in the appendix.

\section{Conclusion}

We present Evoke, a three-step video world model for long-horizon interactive generation.
Evoke decouples persistent world state from ever-growing model context through external
geometric memory, while a teacher redesigned for long-horizon generation provides extended
supervision for the few-step student. This design enables continuous camera and text control
while keeping the state and context of each recurrent generation step bounded as the session
grows. Experiments show that Evoke achieves competitive performance on interactive
world-model benchmarks while sustaining stable hour-scale generation with bounded
computational cost.

Several directions remain open. First, the current geometric world state primarily preserves
coarse scene structure, while fine-grained consistency of object identity, appearance, and
local details remains limited. Richer object-level or semantic world representations may
provide stronger long-term identity consistency. Second, a persistent world should model not
only static geometry but also dynamic state, including object motion, state transitions, and
their long-term evolution. Developing explicit representations that can continuously update
such dynamic world state is an important next step. Finally, further inference acceleration
remains necessary for truly real-time interaction, including higher-compression video VAEs,
more efficient few-step generators, and lower-cost geometric conditioning.

\clearpage

\newpage

\beginappendix
\section{Training curves}
\label{app:training}
\begin{figure}[H]
  \centering
  \includegraphics[width=\linewidth]{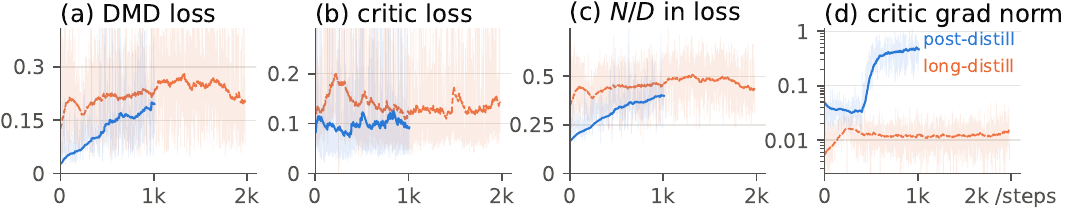}
  \vspace{-5mm}
  \caption{\textbf{Both distillation stages train stably.} \emph{Long-distill} (Evoke Teacher,
  $6\times8$ GPUs, $1981$ steps) and the short \emph{post-distill} continuation that
  yields the released student; faint per-step, bold $120$-step exponential moving average.
  \textbf{(a)} DMD stays bounded in both. \textbf{(b)} The critic converges.
  \textbf{(c)} The normalized gradient entering the loss plateaus; we release at its onset.
  \textbf{(d)} Gradient norms stay bounded across $48$ GPUs and eight scheduler restarts.}
  \label{fig:c}
\vspace{-4mm}
\end{figure}

Figure~\ref{fig:c} is the optimization record of the two distillation stages that
produce the released Evoke student: \emph{long-distill} against the Evoke Teacher
on $6\times8$ GPUs, and the short \emph{post-distill} continuation that yields the
released student.

%
%

\section{Cost of the geometric path}
\label{app:warpcost}
\begin{figure}[H]
  \centering
  \includegraphics[width=\linewidth]{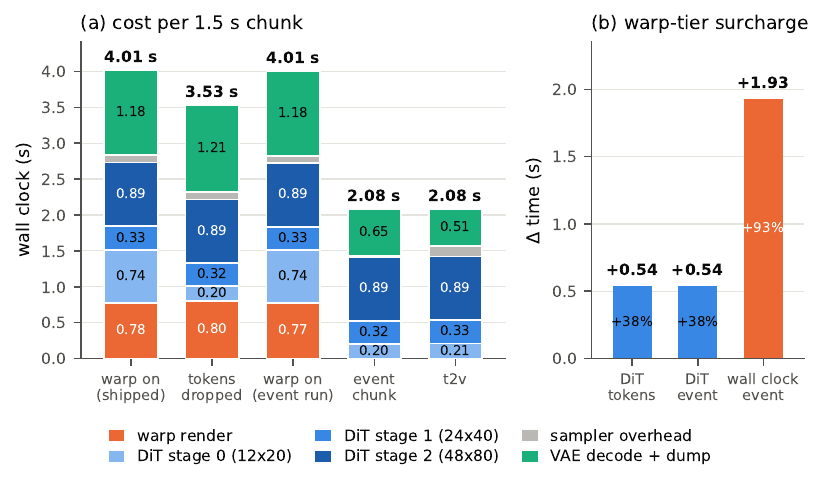}
  \vspace{-5mm}
  \caption{\textbf{The geometric path costs $38\%$ of the denoiser and $93\%$ of a recurrent step that skips
  it.} Per-chunk wall clock on one H200 at $384\!\times\!640$, 3 denoising steps, CFG off; arms differ
  only by the named flag. \textbf{(a)} Warp rendering and coarsest-stage token admission
  are separable costs, and stages~1 and~2 are unchanged throughout---tokens, not geometry,
  carry denoiser cost. \textbf{(b)} Geometry adds $1.84$\,s per chunk, scaling with warp
  coverage: unobserved space costs less exactly where memory offers less.}
  \label{fig:e}
\vspace{-4mm}
\end{figure}

Figure~\ref{fig:e} breaks one recurrent step's wall clock into the denoiser and the geometric
path, and isolates the coarsest-stage token surcharge that \S\ref{sec:runtime}
attributes to warp conditioning. It is also the source of the coverage fit reported
there and in \S\ref{sec:exp-efficiency}.

%
%
%

\section{Long-Horizon Generation and Teacher Ablations}
\label{app:long-horizon-analysis}

The long-session evaluation in Sec.~\ref{sec:exp-session} consists of eight continuous
rollouts, each spanning $2{,}619$ recurrent steps, $65.5$ minutes, and $94{,}281$ generated
frames. All runs use three CFG-free evaluations per chunk at $384\times640$ on a single
H200, with the world state bank restricted to $90$\,s of retained observations. The reported
$2.11$\,s latency refers to diffusion wall clock rather than end-to-end execution. A further
hour-scale rollout was run as a consistency check; its photometric and content-similarity
trajectories remain within the range of the eight-session evaluation.

Long-session behavior is evaluated along complementary photometric and content-based
measures. Photometric statistics quantify progressive changes in brightness, saturation,
and related low-level appearance, while the content descriptor measures similarity to each
session's own opening segment. The latter is a self-consistency measure rather than a
reference-based identity metric. Real-video controls exhibit substantial descriptor
decorrelation as the camera naturally traverses new content, so absolute similarity values
should not be interpreted as permanent preservation of scene identity. The relevant
observation is instead whether generated trajectories continue to diverge progressively with
elapsed time. Full per-session curves and the corresponding real-video controls are provided
here to complement the aggregated results in the main paper.

To isolate the effect of long-horizon teacher supervision, we additionally compare few-step
students distilled with matched recipes from the short-horizon teacher and the Evoke
teacher. The long-horizon teacher yields a clear improvement in photometric stability over
extended rollouts, while the measured content descriptor does not significantly separate the
two students. Sharpness does not improve under the same comparison, and the final continued
checkpoint shows no measured drift advantage over the long-horizon distilled checkpoint.
These results therefore localize the demonstrated transfer benefit to photometric stability
rather than implying a uniform improvement across all measures of long-range consistency.

We further examine whether this benefit can be reduced to a simple dependence on the
scoring-window length $W$. Across $13$ controlled perturbation conditions, teacher-critic
detectability changes little once the window exceeds $W=2$ chunks, and no consistent sharp
threshold emerges as $W$ increases. Varying the temporal distance between the perturbation
and the scored chunk produces similar response curves, indicating that a simple
window-coverage explanation is insufficient for the observed teacher advantage. Analysis
over $5{,}749$ logged training steps further shows that stochastic variation from individual
teacher-critic evaluations is large relative to the systematic drift component, although
aggregation across evaluations can recover the latter. Thus, the empirical benefit of the
Evoke Teacher should be attributed to the trained long-horizon supervision process as a
whole rather than to a single detectability threshold determined by $W$.

Finally, the timed-conditioning ablation separates the control interface from its causal
attribution. Mid-session prompt changes are supported by the per-chunk conditioning
representation, but matched short- and long-horizon teacher variants do not significantly
differ in event-realization rate at the current sample size. The main paper therefore treats
per-chunk conditioning as the mechanism by which timed instructions are represented, without
attributing the observed evocation capability solely to the teacher horizon or conditioning
schedule.

\begin{figure}[p]
  \centering
  \includegraphics[width=5.5in]{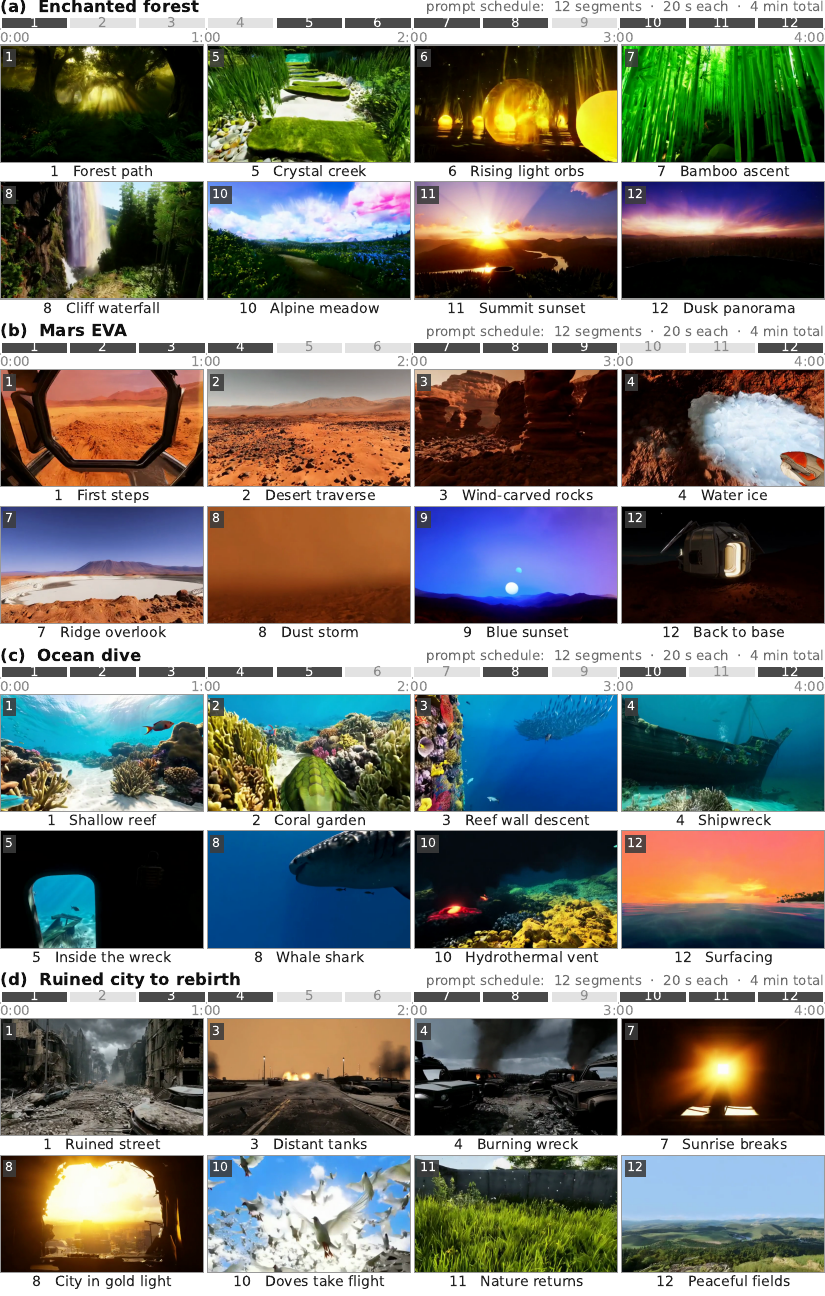}
  \caption{\textbf{Four-minute Evoke Teacher rollouts under a twelve-prompt schedule.} Four sessions~(a--d) generated by the Evoke Teacher rather than the three-step student, each driven by twelve instructions held for $20$\,s apiece. The grey strip above each block is that schedule on a common time axis: the eight segments illustrated below are filled dark, the four that are not remain pale. Each tile is one frame selected inside its own segment rather than sampled at a fixed offset.}
  \label{fig:o}
\vspace{-4mm}
\end{figure}

Figure~\ref{fig:o} shows that interface at the Evoke Teacher's own horizon rather than the
student's: four four-minute rollouts generated by the teacher, each driven by twelve
consecutive instructions held for $20$\,s. Every instruction takes visible effect inside its own segment and the session
runs across all twelve without a reset. The tiles are selected within their segments rather
than sampled at a fixed offset, as the caption states, so the figure documents the behavior
of the conditioning schedule and carries no measurement.

\section{Qualitative rollouts}
\label{app:showcase}
\begin{figure}[p]
  \centering
  \includegraphics[width=\linewidth]{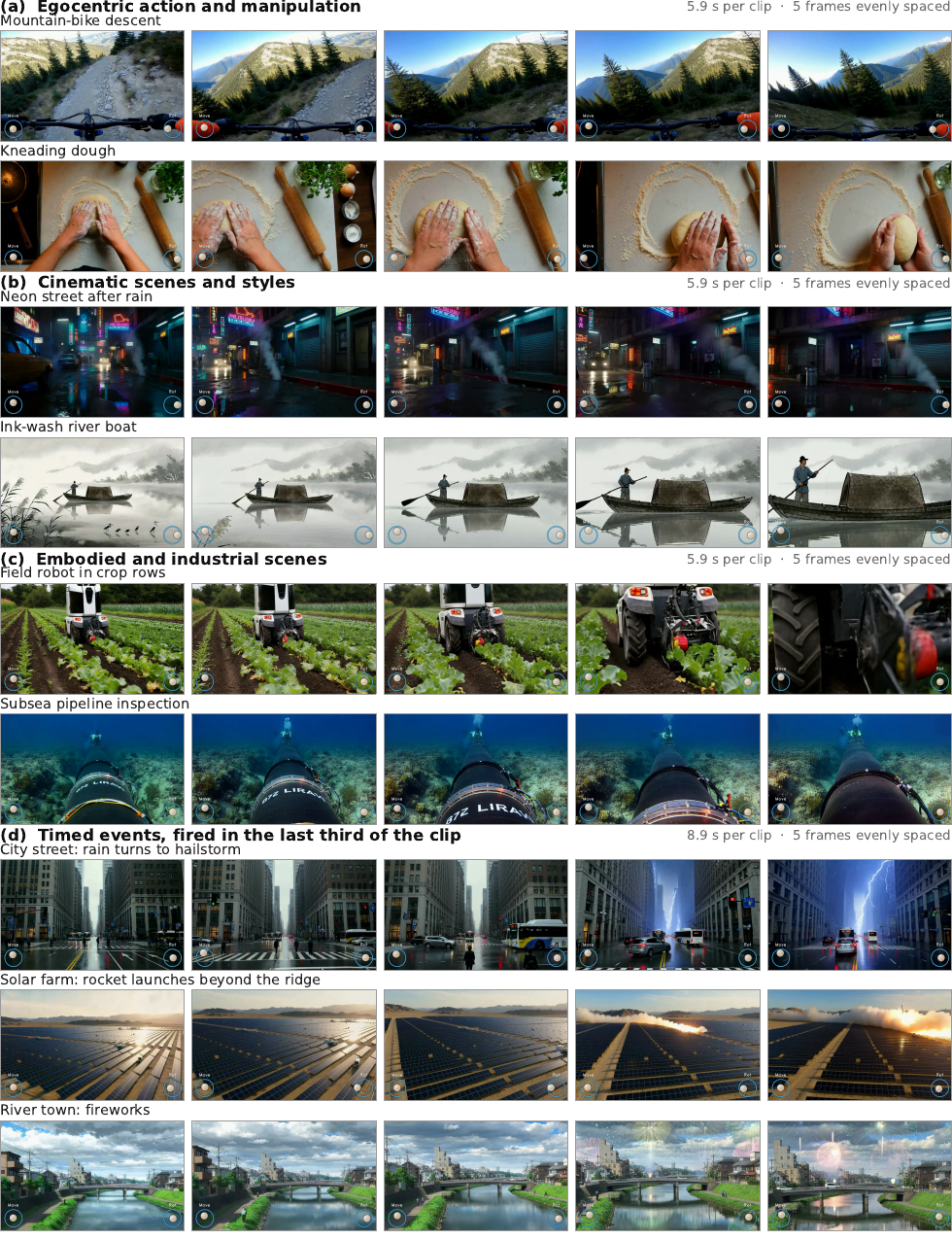}
  \caption{\textbf{Qualitative rollouts across four domains.} Nine sessions generated by Evoke in three steps per chunk without classifier-free guidance, one per row, each shown as five frames evenly spaced over the clip: $5.9$\,s in~(a--c) and $8.9$\,s in~(d), where a timed event fires in the last third. The rings in the lower corners are the engine's own joystick overlay.}
  \label{fig:p}
\vspace{-4mm}
\end{figure}

Figure~\ref{fig:p} collects nine sessions across four settings: egocentric action and
manipulation, cinematic scenes and styles, embodied and industrial scenes, and timed events.
Each row samples five frames evenly along its own clip rather than at chosen moments, so a
row shows what the rollout does over its whole length; the event rows are longer because the
timed instruction fires in the last third. The rings in the lower corners of every tile are
the interactive demo's joystick overlay, which shows the camera command driving the
motion. Like Fig.~\ref{fig:o}, the sheet is qualitative
and carries no measurement.

\section{Public leaderboards in full}
\label{app:vbench}
Table~\ref{tab:vbench} in the main text summarizes two public leaderboards~\citep{vbench2,vbenchpp};
Tables~\ref{tab:vbench2} and~\ref{tab:vbench-long} give the full top-ten of each,
and Tables~\ref{tab:vbench2-dims} and~\ref{tab:vbench-long-dims} the per-dimension
scores behind them. Peer rows are the public leaderboard as of 2026-08-09. Both
breakdowns are reported at the scale on which columns are comparable across rows:
VBench-2.0 aggregates raw dimension scores without normalization, whereas
VBench-Long min--max-normalizes before weighting, so the Evoke row of
Table~\ref{tab:vbench-long-dims} is de-normalized back to raw with the official
constants.

\paragraph{Protocol deviations.} Our VBench-2.0 row departs from the official
protocol in four declared ways: one
sample per prompt rather than five, with \emph{Diversity} keeping the $20$ its
metric requires; $5.875$\,s clips; $640\times384$ resolution; and prompt
augmentation, which the official protocol permits and which leaves case selection,
dimension membership and every answer key byte-identical to the official metadata.
Our VBench-Long row departs in three ways: one sample per prompt;
$8.875$\,s clips rather than $10$\,s, which the official code supports and which two
controlled probes put at $\approx\!+0.02$ Total in our favor; and the same prompt
augmentation. Single-sampling inflates variance but not the expectation, and the
$0.05$ margins reported in Table~\ref{tab:vbench} are of that order, so neither
ranking should be over-read.


\begin{table}[H]
\caption{VBench-2.0, top-10 of the public leaderboard, Evoke
included. All scores $\in[0,100]$, higher is better; \textbf{bold} is best,
\underline{underline} second best.}
\label{tab:vbench2}
\centering
\renewcommand{\arraystretch}{1.05}
\setlength{\tabcolsep}{4pt}
\footnotesize
\begin{tabular}{@{}l c ccccc@{}}
\toprule
\textbf{Model} & \textbf{Total} & \textbf{Creat.} & \textbf{Comm.} & \textbf{Contr.} & \textbf{Human} & \textbf{Phys.} \\
\midrule
\textbf{Evoke (3 step)} & \textbf{66.77} & 55.22 & \textbf{76.62} & 41.99 & \textbf{94.23} & 65.80 \\
Veo 3 & \underline{66.72} & \textbf{60.85} & 69.48 & \textbf{47.04} & 86.88 & 69.35 \\
JT-CV & 64.60 & 51.80 & 66.94 & \underline{45.47} & 83.53 & \textbf{75.23} \\
ABot-World v0.1 & 64.47 & 55.29 & \underline{72.93} & 44.13 & 85.36 & 64.61 \\
Vidu Q1 (2025-04-17) & 62.70 & 56.54 & 65.98 & 38.13 & 81.24 & \underline{71.63} \\
ToMoviee 2.0 & 61.78 & 45.96 & 67.41 & 45.37 & 80.68 & 69.47 \\
Wan2.1 & 60.20 & 55.25 & 63.98 & 37.32 & 81.60 & 62.84 \\
Seedance 1.0 Pro (2025-05-28) & 59.81 & 53.04 & 64.31 & 39.84 & 77.06 & 64.81 \\
Kling 1.6 & 59.00 & 48.58 & 65.45 & 33.05 & 83.56 & 64.35 \\
Sora-480p & 58.38 & \underline{60.57} & 64.32 & 22.09 & \underline{87.72} & 57.18 \\
\bottomrule
\end{tabular}

\end{table}


\begin{table}[H]
\caption{VBench-Long, top-10 of the public leaderboard, Evoke
included. All scores $\in[0,100]$, higher is better; \textbf{bold} is best,
\underline{underline} second best.}
\label{tab:vbench-long}
\centering
\renewcommand{\arraystretch}{1.05}
\setlength{\tabcolsep}{5pt}
\small
\begin{tabular}{@{}l ccc@{}}
\toprule
\textbf{Model} & \textbf{Total} & \textbf{Quality} & \textbf{Semantic} \\
\midrule
IPOW & \textbf{88.26} & \textbf{87.83} & \textbf{90.01} \\
Vidu Q1 (2025-04-17) & \underline{87.41} & \underline{87.28} & \underline{87.94} \\
IPOC (2025-04-14) & 86.57 & 87.00 & 84.84 \\
Wan2.1(2025-02-24) & 86.22 & 86.67 & 84.44 \\
IPOC & 85.71 & 86.12 & 84.09 \\
MiracleVision V5 & 85.23 & 86.68 & 79.43 \\
\textbf{Evoke (3 step)} & 85.11 & 85.55 & 83.36 \\
Veo 3 & 85.06 & 85.70 & 82.49 \\
LanDiff & 84.87 & 85.41 & 82.72 \\
Wan2.1 & 84.70 & 85.64 & 80.95 \\
\bottomrule
\end{tabular}

\end{table}


\begin{table}[H]
\caption{Per-dimension VBench-2.0 breakdown for the systems of
Table~\ref{tab:vbench2}. Raw scores in $[0,100]$, higher is better;
\textbf{bold} is best, \underline{underline} second best.}
\label{tab:vbench2-dims}
\centering
\renewcommand{\arraystretch}{1.05}
\setlength{\tabcolsep}{2.5pt}
\tiny
\resizebox{\ifdim\width>\linewidth\linewidth\else\width\fi}{!}{%
\begin{tabular}{@{}l *{18}{c}@{}}
\toprule
& \multicolumn{2}{c}{\textbf{Creativity}} & \multicolumn{2}{c}{\textbf{Commonsense}} & \multicolumn{7}{c}{\textbf{Controllability}} & \multicolumn{3}{c}{\textbf{Human Fidelity}} & \multicolumn{4}{c}{\textbf{Physics}} \\
\cmidrule(lr){2-3}\cmidrule(lr){4-5}\cmidrule(lr){6-12}\cmidrule(lr){13-15}\cmidrule(lr){16-19}
\textbf{Model} & \rotatebox{90}{\textbf{Composition}} & \rotatebox{90}{\textbf{Diversity}} & \rotatebox{90}{\textbf{Instance Preserv.}} & \rotatebox{90}{\textbf{Motion Rational.}} & \rotatebox{90}{\textbf{Dyn. Spatial Rel.}} & \rotatebox{90}{\textbf{Dyn. Attribute}} & \rotatebox{90}{\textbf{Motion Order}} & \rotatebox{90}{\textbf{Human Interaction}} & \rotatebox{90}{\textbf{Complex Landscape}} & \rotatebox{90}{\textbf{Complex Plot}} & \rotatebox{90}{\textbf{Camera Motion}} & \rotatebox{90}{\textbf{Human Anatomy}} & \rotatebox{90}{\textbf{Human Identity}} & \rotatebox{90}{\textbf{Human Clothes}} & \rotatebox{90}{\textbf{Mechanics}} & \rotatebox{90}{\textbf{Thermotics}} & \rotatebox{90}{\textbf{Material}} & \rotatebox{90}{\textbf{Multi-View Cons.}} \\
\midrule
\textbf{Evoke (3 step)} & \textbf{70.38} & 40.07 & 87.72 & \textbf{65.52} & 36.23 & 50.55 & 35.35 & 71.00 & 18.00 & 12.42 & \textbf{70.37} & 89.23 & \textbf{93.45} & \textbf{100.00} & 80.00 & \underline{75.00} & 83.33 & 24.87 \\
Veo 3 & \underline{68.57} & 53.13 & \underline{92.98} & 45.98 & \textbf{45.89} & \textbf{63.74} & \textbf{40.40} & \underline{80.67} & \textbf{21.78} & \textbf{21.87} & 54.94 & 90.26 & 70.84 & \underline{99.53} & \textbf{81.82} & \textbf{75.51} & \underline{84.11} & 35.95 \\
JT-CV & 58.75 & 44.86 & 73.53 & \underline{60.34} & \underline{44.05} & 54.21 & 35.35 & \textbf{84.67} & 19.11 & 14.87 & 66.05 & \textbf{93.64} & 64.51 & 92.44 & 76.69 & 66.21 & 81.25 & \textbf{76.78} \\
ABot-World v0.1 & 55.06 & 55.52 & \underline{92.98} & 52.87 & 31.88 & \underline{61.90} & \underline{39.39} & \underline{80.67} & \underline{21.33} & 12.01 & 61.73 & \underline{91.77} & 66.27 & 98.05 & 68.42 & 73.97 & \textbf{85.34} & 30.72 \\
Vidu Q1 (2025-04-17) & 60.94 & 52.14 & 86.55 & 45.40 & 15.46 & 55.68 & 34.01 & 73.67 & 17.56 & 10.67 & 59.88 & 88.87 & 73.31 & 81.54 & \underline{81.75} & 64.03 & 72.82 & \underline{67.92} \\
ToMoviee 2.0 & 51.78 & 40.13 & 85.96 & 48.85 & 40.10 & 61.17 & 37.71 & 75.33 & 17.78 & \underline{15.46} & \underline{70.06} & 76.78 & 75.16 & 90.10 & 74.24 & 67.35 & 82.46 & 53.84 \\
Wan2.1 & 54.51 & \underline{56.00} & 88.89 & 39.08 & 31.40 & 45.42 & 35.35 & 80.00 & 16.44 & 13.14 & 39.51 & 82.59 & 65.73 & 96.48 & 71.74 & 67.16 & 75.68 & 36.79 \\
Seedance 1.0 Pro (2025-05-28) & 51.91 & 54.18 & 80.92 & 47.70 & 37.20 & 44.32 & 29.63 & 79.67 & 16.67 & 13.98 & 57.41 & 64.05 & 67.13 & \textbf{100.00} & 63.16 & 64.34 & 77.38 & 54.37 \\
Kling 1.6 & 43.89 & 53.26 & 92.40 & 38.51 & 20.77 & 19.41 & 29.29 & 72.00 & 17.33 & 10.83 & 61.73 & 86.99 & 71.95 & 91.75 & 65.55 & 59.46 & 68.00 & 64.38 \\
Sora-480p & 53.65 & \textbf{67.48} & \textbf{94.15} & 34.48 & 19.81 & 8.06 & 15.15 & 58.00 & 15.33 & 11.11 & 27.16 & 86.45 & \underline{78.57} & 98.15 & 62.22 & 43.36 & 64.94 & 58.22 \\
\bottomrule
\end{tabular}
}

\vspace{2pt}
\begin{minipage}{\linewidth}\footnotesize
Evoke is sampled in $3$ steps (three-stage pyramid, one step per stage) with no CFG, i.e.\ $3$ network evaluations per chunk, against peers running their own many-step default samplers.
The four protocol deviations of Table~\ref{tab:vbench2} apply unchanged.
\emph{Diversity} alone is generated from the official raw prompts, since
augmenting it was measured to cost $3.38$ points.
\end{minipage}
\end{table}


\begin{table}[H]
\caption{Per-dimension VBench-Long breakdown for the systems of
Table~\ref{tab:vbench-long}. \emph{Raw} scores in $[0,100]$, before the
min--max normalization the aggregates apply; higher is better,
\textbf{bold} is best, \underline{underline} second best.}
\label{tab:vbench-long-dims}
\centering
\renewcommand{\arraystretch}{1.05}
\setlength{\tabcolsep}{2.5pt}
\tiny
\resizebox{\ifdim\width>\linewidth\linewidth\else\width\fi}{!}{%
\begin{tabular}{@{}l *{16}{c}@{}}
\toprule
& \multicolumn{7}{c}{\textbf{Quality dimensions}} & \multicolumn{9}{c}{\textbf{Semantic dimensions}} \\
\cmidrule(lr){2-8}\cmidrule(lr){9-17}
\textbf{Model} & \rotatebox{90}{\textbf{Subject Cons.}} & \rotatebox{90}{\textbf{Background Cons.}} & \rotatebox{90}{\textbf{Temporal Flicker}} & \rotatebox{90}{\textbf{Motion Smooth.}} & \rotatebox{90}{\textbf{Aesthetic}} & \rotatebox{90}{\textbf{Imaging}} & \rotatebox{90}{\textbf{Dynamic Degree}} & \rotatebox{90}{\textbf{Object Class}} & \rotatebox{90}{\textbf{Multiple Objects}} & \rotatebox{90}{\textbf{Human Action}} & \rotatebox{90}{\textbf{Color}} & \rotatebox{90}{\textbf{Spatial Rel.}} & \rotatebox{90}{\textbf{Scene}} & \rotatebox{90}{\textbf{Appearance Style}} & \rotatebox{90}{\textbf{Temporal Style}} & \rotatebox{90}{\textbf{Overall Cons.}} \\
\midrule
IPOW & 97.31 & 97.67 & \underline{99.76} & 97.94 & \textbf{67.42} & 68.86 & \textbf{95.56} & \textbf{97.43} & \textbf{95.96} & 97.00 & \underline{97.15} & \textbf{97.80} & \textbf{76.32} & 23.60 & \textbf{27.59} & 26.77 \\
Vidu Q1 (2025-04-17) & 95.80 & 97.11 & 99.34 & 98.47 & \underline{67.32} & \underline{69.07} & 91.85 & \underline{97.40} & \underline{93.89} & \textbf{99.60} & \textbf{99.04} & 93.74 & \underline{67.41} & 22.37 & \underline{26.46} & 27.24 \\
IPOC (2025-04-14) & 96.60 & 97.57 & \textbf{99.78} & 97.60 & 65.22 & 69.02 & 93.06 & 92.41 & 87.25 & 98.00 & 86.79 & \underline{95.46} & 61.45 & 24.01 & 25.87 & 26.85 \\
Wan2.1(2025-02-24) & 96.62 & 97.58 & 99.38 & 97.42 & 63.45 & \textbf{69.63} & 94.26 & 96.64 & 86.59 & 99.20 & 94.39 & 85.70 & 61.24 & 21.59 & 26.18 & 27.49 \\
IPOC & 96.68 & 97.21 & 99.11 & 97.12 & 64.11 & 68.43 & 92.59 & 94.14 & 85.73 & \underline{99.40} & 88.64 & 90.55 & 54.01 & \underline{24.59} & 25.79 & \underline{27.56} \\
MiracleVision V5 & 97.32 & 96.35 & 98.62 & \underline{99.02} & 65.28 & 65.67 & 93.52 & 93.04 & 81.60 & 98.00 & 89.92 & 76.22 & 46.80 & 21.72 & 25.16 & 26.94 \\
\textbf{Evoke (3 step)} & \textbf{97.92} & \underline{98.10} & 99.52 & 98.96 & 64.43 & 64.94 & 71.39 & 91.65 & 85.67 & 99.00 & 94.76 & 82.82 & 62.15 & 22.25 & 24.70 & 27.31 \\
Veo 3 & \underline{97.36} & 96.89 & 99.30 & \textbf{99.16} & 63.81 & 68.23 & 72.43 & 93.89 & 82.20 & \underline{99.40} & 82.48 & 84.26 & 57.43 & 23.55 & 25.97 & \textbf{27.88} \\
LanDiff & 91.86 & \textbf{98.24} & 99.42 & 97.11 & 65.24 & 65.78 & 93.33 & 94.86 & 87.13 & 96.60 & 91.22 & 73.58 & 54.84 & \textbf{25.60} & 25.28 & 27.43 \\
Wan2.1 & 95.92 & 97.39 & 99.53 & 96.92 & 61.53 & 67.28 & \underline{94.35} & 94.24 & 81.44 & 98.80 & 87.79 & 81.08 & 53.67 & 21.13 & 25.69 & 27.44 \\
\bottomrule
\end{tabular}
}

\vspace{2pt}
\begin{minipage}{\linewidth}\footnotesize
Evoke is sampled in $3$ steps (three-stage pyramid, one step per stage) with no CFG, i.e.\ $3$ network evaluations per chunk, against peers running their own many-step default samplers.
The three protocol deviations of Table~\ref{tab:vbench-long} apply unchanged.
The Evoke row is stored normalized in the source data and is de-normalized
here with the official constants so that it shares the leaderboard's scale.
\end{minipage}
\end{table}

\section{WBench public leaderboard}
\label{app:wbench-board}
Table~\ref{tab:wbench} compares Evoke with eight few-step interactive systems on the
WBench navigation split~\citep{wbench}. Table~\ref{tab:wbench-board} places the same run in the public
WBench leaderboard for that split, which ranks thirty systems of any sampling budget by
the unweighted mean of the five group scores; on this split the Interaction group is the
navigation dimension alone. The comparison is not step-matched: Evoke is sampled in three
steps without classifier-free guidance, whereas most rows around it run their own
many-step defaults. Our row leads the board by $0.1$ Average, a margin of the same order
as the ones \S\ref{sec:exp-setup} declines to read as wins.


\begin{table}[H]
\caption{\textbf{WBench public leaderboard, Navi split} ($n=158$ cases), top ten by Average, Evoke included. Average is the unweighted mean of the five group scores; peer rows are the public leaderboard as of 2026-08-12. All scores $\in[0,100]$, higher is better; \textbf{bold} is best, \underline{underline} second best.}
\label{tab:wbench-board}
\centering
\renewcommand{\arraystretch}{1.05}
\setlength{\tabcolsep}{4pt}
\footnotesize
\begin{tabular}{@{}r l c ccccc@{}}
\toprule
\textbf{\#} & \textbf{Model} & \textbf{Average} & \textbf{Quality} & \textbf{Setting} & \textbf{Inter.} & \textbf{Consist.} & \textbf{Phys.} \\
\midrule
1 & \textbf{Evoke (ours, 3 step)}$^{\dagger}$ & \textbf{80.8} & \textbf{82.8} & 83.8 & 78.6 & 86.9 & \textbf{72.1} \\
2 & HiDream-O1-World & \underline{80.7} & \underline{81.9} & 81.9 & 79.5 & \underline{88.0} & \textbf{72.1} \\
3 & LingBot-World v2 \textit{fast} & 79.4 & 81.8 & 76.8 & 82.8 & 86.5 & 69.1 \\
4 & Kling 3.0 & 79.0 & 81.4 & \underline{91.0} & 69.4 & 83.7 & 69.3 \\
5 & LingBot-World \textit{base-camera} & 78.5 & 78.9 & 72.6 & 80.1 & \textbf{89.9} & 71.2 \\
6 & Wan 2.7 & 78.1 & 81.5 & \textbf{91.4} & 64.4 & 81.6 & \underline{71.8} \\
7 & HY-World 1.5 \textit{ar-distill} & 78.1 & 78.1 & 72.2 & \textbf{86.8} & 86.9 & 66.3 \\
8 & HY-Video 1.5 & 77.9 & 77.6 & 85.6 & 71.4 & 87.4 & 67.4 \\
9 & LingBot-World \textit{fast} & 77.4 & 79.4 & 77.9 & 79.2 & 84.9 & 65.7 \\
10 & Happy Oyster & 76.8 & 77.3 & 74.2 & \underline{84.9} & 84.3 & 63.5 \\
\bottomrule
\end{tabular}

\vspace{2pt}
\begin{minipage}{\linewidth}\footnotesize
$^{\dagger}$~Our own evaluation of the released student, not a leaderboard submission. Evoke is sampled in three steps with no classifier-free guidance against peers running their own default samplers, so the comparison is not step-matched.
\end{minipage}
\end{table}

\end{document}